\documentclass[letterpaper, 10 pt, conference]{ieeeconf}
\pdfoutput=1

\IEEEoverridecommandlockouts                              

\usepackage{graphics} 
\usepackage{subfigure}
\usepackage{epsfig} 
\usepackage{cite} 
\usepackage{paralist}
\usepackage[linesnumbered,vlined,ruled]{algorithm2e}
\SetKw{Continue}{continue}

\usepackage{amsmath}
\usepackage{amssymb}
\usepackage{wasysym}

\newcommand{\props}{\Pi}
\newcommand{\labeling}{\mathcal{L}}
\newcommand{\satisfies}{\models}
\newcommand{\always}{\square}
\newcommand{\eventually}{\Diamond}
\newcommand{\next}{\ocircle}
\newcommand{\until}{\mathcal{U}}

\newcommand{\true}{\relax\ifmmode \mathit{True} \else \em True \/\fi}
\newcommand{\false}{\relax\ifmmode \mathit{False} \else \em False \/\fi}
\newcommand{\aand}{\wedge}
\newcommand{\oor}{\vee}

\newcommand{\states}{S}
\newcommand{\controls}{U}
\newcommand{\trajs}{X}
\newcommand{\sinit}{s_{\mathsf{init}}}
\newcommand{\Sgoal}{\states_{\mathsf{goal}}}
\newcommand{\traj}{x}
\newcommand{\control}{u}
\newcommand{\ftime}{\mathcal{F}}

\newcommand{\siFLTL}{\textrm{si-FLTL}_{\mathsf{G}}}
\newcommand{\siFLTLX}{\textrm{si-FLTL}_{\mathsf{G_X}}}
\newcommand{\vanish}{\mathsf{vanish}}
\newcommand{\spec}{\varphi}
\newcommand{\priority}{\rho}
\newcommand{\pspec}{\mathcal{P}}

\newcommand{\denext}{\mathsf{denext}}

\newcommand{\PX}{P_{\mathsf{X}}}

\newcommand{\automaton}{\mathcal{A}}

\newcommand{\kripke}{\mathcal{K}}
\newcommand{\wkripke}{\overline{\mathcal{K}}}
\newcommand{\kstates}{S_\kripke}
\newcommand{\ksinit}{s_{\mathsf{init}, \kripke}}
\newcommand{\kSgoal}{S_{\mathsf{goal}, \kripke}}
\newcommand{\ksgoal}{s_{\mathsf{goal}, \kripke}}
\newcommand{\krel}{\mathcal{R}_\kripke}
\newcommand{\kprops}{\Pi_\kripke}
\newcommand{\klabeling}{\mathcal{L}_\kripke}
\newcommand{\kdelta}{\Delta_\kripke}
\newcommand{\kweight}{\mathcal{W}_\kripke}
\newcommand{\tword}{\omega}
\newcommand{\word}{w}

\newcommand{\Traces}{\mathsf{Traces}}

\newcommand{\num}[1]{\relax\ifmmode \mathbb #1\else $\mathbb #1$\fi}
\newcommand{\naturals}{{\num N}}
\newcommand{\reals}{{\num R}}
\newcommand{\naturalsle}[1]{\naturals_{\leq #1}}

\newcommand{\union}{\hspace{1mm}\cup\hspace{1mm}}
\newcommand{\intersect}{\hspace{1mm}\cap\hspace{1mm}}

\newcommand{\sample}{\mathsf{sample}}
\newcommand{\steer}{\mathsf{steer}}
\newcommand{\near}{\mathsf{near}}
\newcommand{\tcost}{\mathsf{cost}}
\newcommand{\costtocome}{J_\kripke}

\newcommand{\addstate}{\mathsf{add}}
\newcommand{\connect}{\mathsf{connect}}
\newcommand{\parent}{\mathsf{parent}}

\newcommand{\snew}{s_{\mathsf{new}}}

\newtheorem{lemma}{Lemma}
\newtheorem{proposition}{Proposition}

\newtheorem{definition}{Definition}
\newtheorem{remark}{Remark}
\newtheorem{problem}{Problem}

\newcommand{\footprint}{\mathsf{FP}}
\newcommand{\obs}{\mathsf{SV}}
\newcommand{\clearance}{\mathsf{CZ}}
\newcommand{\lane}{\mathsf{LN}}
\newcommand{\road}{\mathsf{RD}}
\newcommand{\close}{\mathsf{close}}
\newcommand{\onroad}{\mathsf{road}}
\newcommand{\inlane}{\mathsf{lane}}
\newcommand{\collision}{\mathsf{collision}}

\title{\LARGE \bf
  Minimum-Violation Planning for Autonomous Systems:\\
  Theoretical and Practical Considerations
}

\author{Tichakorn Wongpiromsarn$^{1}$, Konstantin Slutsky$^{1}$, Emilio Frazzoli$^{2}$ and Ufuk Topcu$^{3}$
  \thanks{*This work was partially supported by
    AFC Robotics Center of Excellence with award number W911NF1920333.}
  \thanks{$^{1}$Tichakorn Wongpiromsarn and Konstantin Slutsky are with Iowa State University, IA
    {\tt\small nok@iastate.edu, kslutsky@gmail.com}}%
  \thanks{$^{2}$Emilio Frazzoli is with ETH Zurich, Switzerland
    {\tt\small efrazzoli@ethz.ch}}%
  \thanks{$^{3}$Ufuk Topcu is with the University of Texas at Austin, TX
    {\tt\small utopcu@utexas.edu}}%
}

\begin{document}

\maketitle
\thispagestyle{empty}
\pagestyle{empty}

\begin{abstract}

  This paper considers the problem of computing an optimal trajectory
  for an autonomous system that is subject to a set of potentially conflicting rules.
  First, we introduce the concept of prioritized safety specifications,
  where each rule is expressed as a temporal logic formula with its associated
  weight and priority.
  The optimality is defined based on the violation of such prioritized safety specifications.
  We then introduce a class of temporal logic formulas called $\siFLTLX$
  and develop an efficient, incremental sampling-based approach to solve this
  minimum-violation planning problem with guarantees on asymptotic optimality.
  We illustrate the application of the proposed approach in autonomous vehicles,
  showing that $\siFLTLX$ formulas are sufficiently expressive to describe many traffic rules.
  Finally, we discuss practical considerations and present simulation results for a vehicle
  overtaking scenario.

\end{abstract}

\section{INTRODUCTION}

\begin{figure}[h]
  \centering
  \includegraphics[width=0.49\textwidth,trim={5cm 6.5cm 4.5cm 8cm},clip]{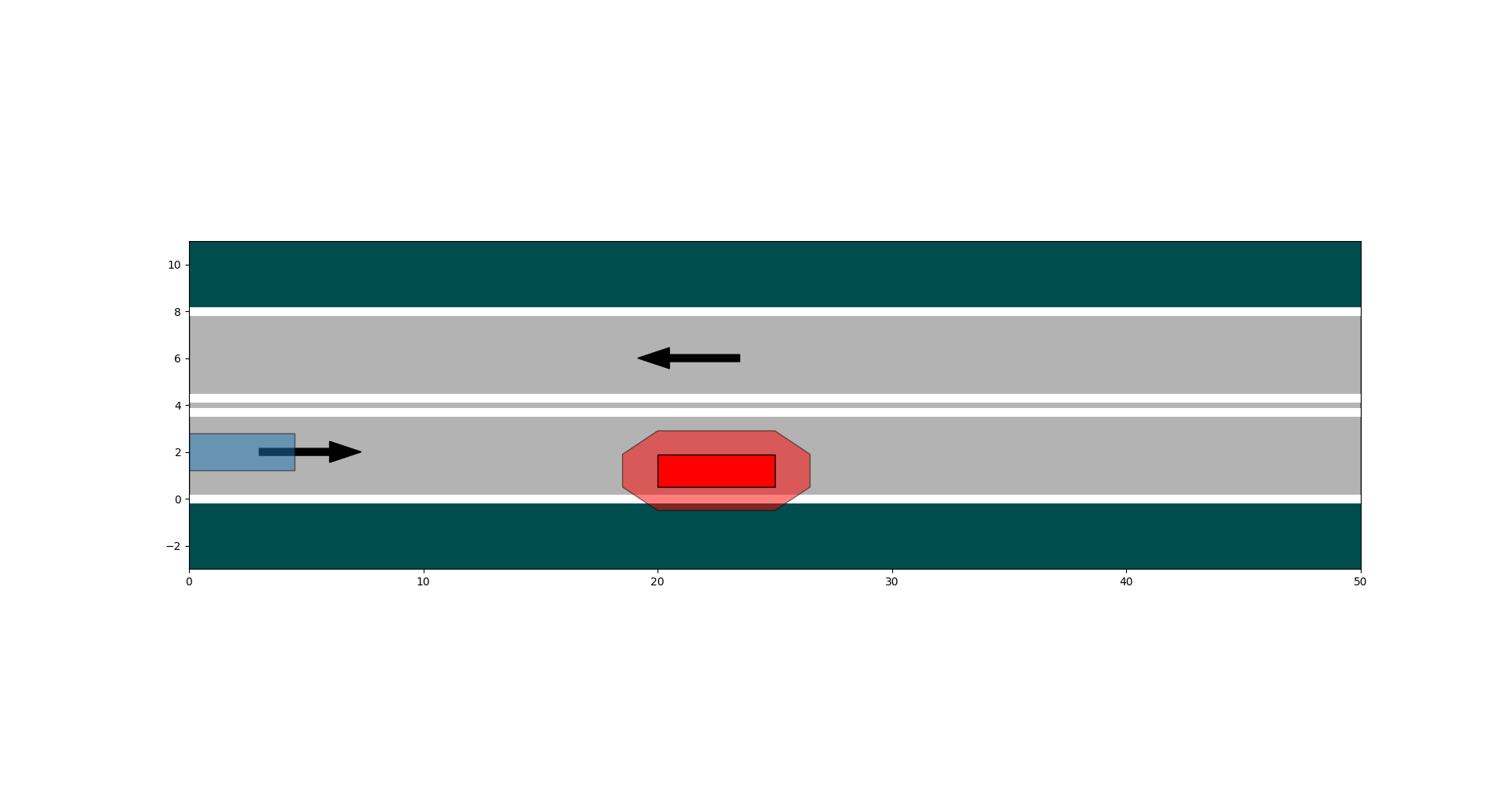}
  \vspace{-8mm}
  \caption{
    The autonomous vehicle (blue rectangle) encounters a stationary vehicle (red rectangle)
    on a two-lane road with a double white lane divider.
    The red octagon represents the clearance zone around the stationary vehicle.
  }
  \vspace{-6mm}
  \label{fig:exp-setup}
\end{figure}

Autonomous vehicles are subject to several road rules.
Often, these rules cannot be simultaneously satisfied.
For example, item 221 of Singapore's \textit{Final Theory of Driving}~\cite{SG-FTD}
suggests keeping a safe gap of one meter when passing by a parked vehicle,
while item 52 of Singapore's \textit{Basic Theory of Driving} \cite{SG-BTD}
prohibits crossing a solid double white lane divider.
As a result, when encountering a vehicle that is improperly parked in a lane with
a solid double white lane divider
as shown in Figure~\ref{fig:exp-setup},
an autonomous vehicle may need to violate either of the aforementioned rules
unless the lane is wide enough to laterally accommodate two cars with a buffer of one meter.

Previous work shows that linear temporal logic (LTL) \cite{Baier:2008:Principles} is
a powerful  language  for  specifying  complex
properties such as traffic rules
\cite{Kress-Gazit:2008:RAM,Kress-Gazit:2008:CASE,Wongpiromsarn:2011:ITSC}.
Existing controller synthesis algorithms for LTL specifications include
closed system synthesis \cite{Kloetzer:2008:Fully},
reactive synthesis \cite{Wongpiromsarn:2012:TAC,Kress-Gazit:2008:CASE,Kress-Gazit:2011:RAM},
and probabilistic synthesis \cite{Wongpiromsarn:2012:CDC,Ulusoy:2014:IJRR,Lahijanian:2012:TR}.
Closed system synthesis considers a deterministic system
and generates a controller, if one exists, for the system to satisfy the specification.
In contrast, reactive synthesis considers a system operating in a nondeterministic,
adversarial environment and ensures that the system satisfies the specification for
all possible adversarial actions.
Finally, probabilistic synthesis considers a probabilistic system and
maximizes the probability that the system satisfies the specification.

More recently, minimum-violation planning
has been proposed to handle conflicting objectives
\cite{Tumova:2013:ACC,Tumova:2013:LCS,Castro:2013:CDC}.
As opposed to reactive and probabilistic synthesis,
minimum-violation planning considers a deterministic system and
relies on real-time re-planning to respond to quickly changing environments.
It is particularly suitable for applications such as autonomous vehicles,
where
(1) it is hard to obtain an accurate probabilistic model of the environment,
(2) accounting for all possible adversarial actions of the environment
may render the system too conservative, and
(3) the system is subject to multiple rules of different importance
and there may be situations where not all the rules can be simultaneously satisfied.
Existing work on minimum-violation planning, however, relies on
converting an LTL specification to a finite automaton,
whose size is exponential in the length of the specification,
making real-time re-planning unrealistic.

As shown in \cite{Wongpiromsarn:2011:ITSC}, most rules of the road
can be expressed by a safety formula.
In particular, this paper considers the case where autonomous vehicles
need to reach the target location but may violate some road rules if needed.
We assume that each rule has a certain penalty associated with its violation.
The goal of motion planning is to minimize such penalties.
The main focus of the paper is on reducing the computational complexity
of minimum-violation planning to make real-time re-planning possible.

This paper is closely related to the framework proposed in \cite{Censi:2019:ICRA}.
The main focus of \cite{Censi:2019:ICRA} is the framework for describing
rules and their associated violation penalties,
taking into various considerations, including
law, ethics, local driving culture, etc.
Given all the rules and their associated violation penalties,
this paper focuses on computing a trajectory of an autonomous vehicle
that minimizes the total penalty.

The main contributions of this paper are twofold.
First, we introduce a class of linear temporal logic formulas
called $\siFLTLX$ that is sufficient to precisely describe many traffic rules.
Second, we propose an efficient algorithm for computing
a path that minimizes the amount of rule violation
with the same computational complexity as
traditional motion planning algorithms.
The proposed algorithm eliminates the exponential part of the complexity of existing algorithms
by avoiding the conversion of specifications into finite automata.
The remainder of the paper is organized as follows:
Section \ref{sec:prelim} introduces the
terminology and notations used throughout the paper.
Section \ref{sec:problem} formulates the minimum-violation planning problem.
Section \ref{sec:theoretical} and Section \ref{sec:practical} discuss
the solution and practical considerations, respectively.
Finally, Section \ref{sec:results} presents
simulation results.


\section{PRELIMINARIES}
\label{sec:prelim}

We consider time-invariant dynamical systems and
use finite linear temporal logic (FLTL) \cite{Manna:1995:TVR}
to specify their correct behaviors.
Given a natural number $n$, let $\naturalsle{n} = \{0, 1, \ldots, n\}$
be the set of natural numbers not greater than $n$.
For any set $\mathcal{S}$, $|\mathcal{S}|$ and $2^{\mathcal{S}}$ denote the cardinality and the powerset of $\mathcal{S}$,
respectively.


\subsection{Systems}
Let $\states \subset \reals^d$, where $d \in \naturals$,
be a compact set of states and
$\controls$ be a compact set of control signals.
Additionally, let $\props$ denote a finite set of atomic propositions
that capture the properties of interest.
We define the labeling function
$\labeling: \states \to 2^{\props}$, which maps each state to
a set of atomic propositions that are true at that state.

%
Consider a time-invariant dynamical system
\begin{equation}
        \dot{\traj}(t) = f(\traj(t), \control(t)),
        \label{eq:dynamical_system}
\end{equation}
with the initial state $\traj(0) = \sinit \in \states$.
$f: \states \times \controls \to \reals^d$ is assumed to be Lipschitz continuous
in both arguments.

Given $T \in \reals_{\geq 0}$, $\traj: [0,T] \to \states$
is a \emph{trajectory} of (\ref{eq:dynamical_system}) if there exists
$\control : [0, T] \to \controls$ satisfying (\ref{eq:dynamical_system})
for all $t \in [0, T]$.
Note that a trajectory does not necessarily start at $\sinit$.
Let $\ftime(\traj) = T$ denote the final time of $\traj: [0, T] \to \states$.

Consider a trajectory $\traj : [0, T] \to \states$.
A \emph{finite timed word} of $\traj$ with respect to a finite set
$\Gamma = \{t_{1}, t_{2}, \ldots, t_{n}\}$
with $0 < t_{1} < t_{2} < \cdots <  t_{n } < T$
is a finite sequence
$\tword(\traj, \Gamma) = l^{t}_0 l^{t}_1 \ldots l^{t}_{n}$
where $l^{t}_i = \bigl(\lim_{t \to t_{i}^+} \labeling(\traj(t)), t_{i+1} - t_i\bigr)$
for all $i \in \naturalsle{n}$, $t_{0} = 0$ and $t_{n+1} = T$.
%
Let $\mathcal{T}(\traj) = \big\{t \in (0, T) \hspace{1mm}|\hspace{1mm}
\lim_{t' \to t^-} \labeling(\traj(t')) \not= \lim_{t' \to t^+} \labeling(\traj(t')\big\}$
be the set of discontinuities of $\labeling(\traj(\cdot))$.
Throughout the paper, we assume that the labeling function $\labeling$
is such that the limits in the definition of $\mathcal{T}(\traj)$ exist and
$\mathcal{T}(\traj)$ is finite,
for any trajectory $x$ of (\ref{eq:dynamical_system}).
The \emph{finite timed word} of $x$ is defined as
$\tword(x) = \tword(x, \mathcal{T}(x))$.

A dynamical system in (\ref{eq:dynamical_system})
can be abstracted into a finite state system represented by
a durational Kripke Structure.

\begin{definition}[Durational Kripke Structure]
A \emph{durational Kripke structure} is a tuple
\begin{equation}
        \kripke = (\kstates, \ksinit, \krel, \kprops, \klabeling, \kdelta),
        \label{eq:kripke}
\end{equation}
where
$\kstates$ is a finite set of states,
$\ksinit \in \kstates$ is the initial state,
$\krel \subseteq \kstates \times \kstates$ is a transition relation,
$\kprops$ is a set of atomic propositions,
$\klabeling: \kstates \to 2^{\kprops}$ is a state labeling function,
and
$\kdelta: \krel \to \reals_{\geq 0}$ is a function assigning a time duration to each transition.
\label{def:kripke}
\end{definition}

A \emph{finite trace} of $\kripke$ is a finite sequence of states
$\tau = s_0 s_1 \ldots s_n$ such that
$s_0 = \ksinit$ and $(s_i, s_{i+1}) \in \krel$,
for all $i \in \naturalsle{n-1}$.
%



\subsection{Specifications}
We consider specifications that can be described by
a subset of FLTL.
Roughly, an FLTL formula is built up from
\begin{inparaenum}[(a)]
\item a set of atomic propositions,
\item the logic connectives:
negation ($\neg$),
disjunction ($\oor$),
conjunction ($\aand$)
and
material implication ($\implies$),
and
\item the temporal operators:
next ($\next$),
always ($\always$),
eventually ($\eventually$)
and until ($\until$).
\end{inparaenum}
We refer to an FLTL formula that does not include temporal operators as
a \emph{propositional logic} formula.

An FLTL formula $\spec$ over $\props$ is interpreted over a finite word
$\word = l_0 l_1 \ldots l_n \in (2^\props)^{n+1}$
and we write $\word \satisfies \spec$ if $\word$ satisfies $\spec$.
In particular, consider $p, p' \in \props$.
$\word \satisfies p$ if and only if $p \in l_0$.
$\word \satisfies \always p$
if and only if $p \in l_i$ for all $i \in \naturalsle{n}$.
Consider a more complicated specification that will be used throughout the paper
\begin{equation}
        \spec = \always \big(p \implies (\next p \oor \next p')\big).
        \label{ex:spec}
\end{equation}
In this case, $\word \satisfies \spec$ if and only if
for any $i \in \naturalsle{n-1}$ such that $p \in l_i$,
we have $p \in l_{i+1}$ or $p' \in l_{i+1}$.
The satisfaction of an FLTL formula with respect to a finite timed word
can be defined in a natural way:
Given an FLTL formula $\spec$ over $\props$ and a finite timed word
$\tword = (l_0, d_0) \ldots (l_n, d_n) \in (2^\props \times \reals_{\geq 0})^{n+1}$,
$\tword \satisfies \spec$ if and only if $\word(\tword) = l_0 \ldots l_n \satisfies \spec$.

\begin{definition}[$\siFLTLX$]
An $\siFLTLX$ formula over a set $\props$ of atomic propositions is an FLTL formula
that is stutter-invariant (see below)
and is of the form
\begin{equation}
  \spec = \always \PX,
  \label{eq:siFLTLX}
\end{equation}
where $\PX$ belongs to the smallest set defined inductively by the following rules:
\begin{itemize}
\item $p$ is a formula for all $p \in \props \cup \{\true, \false\}$,
\item $\next p$ is a formula for all $p \in \props  \cup \{\true, \false\}$, and
\item if $\PX^1$ and $\PX^2$ are formulas, then so are
  $\neg \PX^1$,
  $\PX^1 \oor \PX^2$,
  $\PX^1 \aand \PX^2$ and
  $\PX^1 \implies \PX^2$.
\end{itemize}
In other words, $\PX$ is a Boolean combination of propositions from $\props$ and expressions
of the form $\next p$ where $p \in \props$.
\label{def:siFLTLX}
\end{definition}

Roughly, a specification is stutter-invariant if its satisfaction
with respect to any word is not affected
by operations that duplicate some letters or remove some duplicate letters in that word.
For example, consider $\word = l_0 l_1 \ldots l_n$ and
$\word' = l_0 l_1 \ldots l_{i-1} l_i l_i l_{i+1} \ldots l_n$, which is constructed from
$\word$ by duplicating $l_i$ for some $i \in \naturalsle{n}$.
If $\spec$ is stutter-invariant,
then $\word \satisfies \spec$ if and only if $\word' \satisfies \spec$.
We refer the reader to \cite{Peled:1997:IPL} for the definition
of stutter-invariant properties.
See, e.g., \cite{Klein:2007:CIAA,Michaud:2015:PSC} for approaches to check
whether a specification is stutter-invariant.

Regardless of its simplicity, $\siFLTLX$ turns out to be sufficiently expressive
to describe many traffic rules.
Reference~\cite{Wongpiromsarn:2011:Synthesis} shows
that all the rules enforced in the DARPA Urban Challenge 2007
can be expressed with $\siFLTLX$ formulas.
All the traffic rules in the examples presented in \cite{Castro:2013:CDC}
can also be described using $\siFLTLX$ formulas.

\begin{definition}[Prioritized Safety Specification]
  A \emph{prioritized safety specification} is a tuple
  $\pspec = (\props, \Phi, \Psi, \priority)$ where
  $\props$ is a set of atomic propositions,
  $\Phi$ is a set of $\siFLTLX$ formulas over $\props$,
  $\Psi = (\Psi_0, \Psi_1, \ldots, \Psi_N)$
  organizes the formulas in $\Phi$ into a hierarchy based on their priorities
  such that $\Psi_i \subseteq \Phi$, for all $i \in \naturalsle{N}$,
  and
  $\priority : \Phi \to \naturals$
  is a function that assigns the weight to each
  $\spec \in \Phi$.
  Throughout the paper, we refer to each $\spec \in \Phi$
  as an \emph{atomic safety rule}.
  \label{def:prioritized-safety-specification}
\end{definition}

We use the \textit{level of unsafety} to measure the violation of an
$\siFLTLX$ formula.
%
Consider an $\siFLTLX$ formula $\spec = \always \PX$ and
a finite timed word
$\tword = l^t_0 l^t_1 \ldots l^t_n$ where
$l^t_i = (l_i, d_i) \in 2^\props \times \reals_{\geq 0}$
for all $i \in \naturalsle{n}$.
We let $l^{t}_{n+1} = (l_n,0)$ and
define the level of unsafety of $\tword$ with respect to $\spec$ as
\begin{equation}
  \lambda(\tword, \spec) =
  \mkern-20mu
  \sum_{i \in \naturalsle{n} \hspace{1mm}\big|\hspace{1mm} l_i l_{i+1} \not\satisfies \PX}
  \mkern-20mu
  \tilde{\lambda}(l^{t}_{i}, \PX),
  \label{eq:unsafety-individual-siFLTLX}
\end{equation}
where $\tilde{\lambda}(l^{t}_{i}, \PX) = d_{i}$ if $l_{i}l' \not\satisfies \PX$ for all
$l' \in 2^{\props}$; otherwise $\tilde{\lambda}(l^{t}_{i}, \PX) = 1$.
%
%
Note that this choice of $\tilde{\lambda}$ differentiates
the violation caused by visiting a (unsafe) state with label $l_{i}$
(the case where $l_i l' \not\satisfies \PX$ for all $l' \in 2^{\props}$)
and the violation caused by taking a (unsafe) transition from a state with label $l_{i}$
to a state with label $l_{i+1}$
(the case where
$l_i l_{i+1} \not\satisfies \PX$ but $l_i l' \satisfies \PX$ for some $l' \in 2^{\props}$).
In particular, the cost of visiting an unsafe state is the time spent on that state,
whereas the cost of taking an unsafe transition is 1.
This choice of violation cost is to better accommodate the notion of next as discussed in
Remark~\ref{remark:unsafety}.

Let $\pspec = (\props, \Phi, \Psi, \priority)$ be a prioritized safety specification
where $\Psi = (\Psi_0, \Psi_1, \ldots, \Psi_N)$.
We define the level of unsafety of $\omega$ with respect to $\pspec$ as
\begin{equation}
  \lambda_{\pspec}(\tword) = (\lambda_{\pspec}(\tword, \Psi_0), \ldots, \lambda_{\pspec}(\tword, \Psi_N))
  \in \reals^{N+1},
  \label{eq:unsafety-prioritized}
\end{equation}
where for each $i \in \naturalsle{N}$,
\begin{equation}
  \lambda_{\pspec}(\tword, \Psi_i) =
  \sum_{\spec \in \Psi_i} \priority(\spec) \lambda(\tword, \spec).
  \label{eq:unsafety-set}
\end{equation}


\begin{remark}
  \label{remark:unsafety}
  In \cite{Castro:2013:CDC},
  the level of unsafety of a finite timed word $\tword$
  with respect to an atomic safety rule $\spec$ is defined as
  \begin{equation}
        \lambda(\tword, \spec) = \min_{I \subseteq \naturalsle{n} |
        \vanish(\tword, I) \satisfies \spec}
        \sum_{i \in I} d_i,
        \label{eq:unsafety-ref}
\end{equation}
where for any given finite sequence $w = l_0 l_1 \ldots l_n$
and a set $I \subseteq \naturalsle{n}$,
$\vanish(w, I)$ is defined as a subsequence of $w$ obtained by
removing all $l_i$, $i \in I$.
This definition is consistent with (\ref{eq:unsafety-individual-siFLTLX}) for the case
where $\spec$ is an invariant property, i.e.,
$\PX$ does not include the next operator.
However, for the case where $\PX$ includes the next operator,
(\ref{eq:unsafety-individual-siFLTLX}) and (\ref{eq:unsafety-ref})
may yield different results.
Our choice of (\ref{eq:unsafety-individual-siFLTLX}) is to better accommodate the notion of next,
which is not handled in \cite{Castro:2013:CDC}.
For example, consider a finite timed word
$\tword = (\{p_0\}, d_0) (\{p_1\}, d_1)$
and an $\siFLTLX$ formula
$\spec = \always(p_0 \implies \next p_0)$
where $\props = \{p_0, p_1\}$.
In this case, we get
$\sum_{i \in \naturalsle{1} \hspace{1mm}\big|\hspace{1mm} l_i l_{i+1} \not\satisfies \PX}
\tilde{\lambda}(l_{i}, \PX) = 1$
regardless of the value of $d_0, d_1$;
thus, the level of unsafety defined in (\ref{eq:unsafety-individual-siFLTLX})
corresponds to the number of unsafe transitions.
As a result, this definition allows us to specify an objective such as
minimizing the number of lane changes.
In contrast,
$\min_{I \subseteq \naturalsle{1} | \vanish(\tword, I) \satisfies \spec} \sum_{i \in I} d_i =
\min(d_0, d_1)$, and thus,
the level of unsafety defined in (\ref{eq:unsafety-ref})
corresponds to the total duration either before or after the unsafe transition.
\end{remark}


%


\section{PROBLEM FORMULATION}
\label{sec:problem}

Consider the time-invariant dynamical system (\ref{eq:dynamical_system})
with the initial state $\sinit$ and the set $\Sgoal \subset \states$
of goal states.
Let
$\trajs = \big\{\traj : [0, T] \to \states \hspace{2mm}\big|\hspace{2mm}
T \in \reals_{\geq 0},
\traj(0) = \sinit,
\traj(T) \in \Sgoal
\big\}$
be the set of trajectories of (\ref{eq:dynamical_system}), 
starting at $\sinit$ and ending at a state $s \in \Sgoal$.
%

Given a prioritized safety specification $\pspec = (\props, \Phi, \Psi, \priority)$,
the minimum-violation planning problem is to compute an optimal trajectory $x^* \in \trajs$
that minimizes the time of reaching a state $s \in \Sgoal$ among all the trajectories
that minimize the level of unsafety with respect to $\pspec$.
Formally, we define the cost function
$J : \trajs \to \reals^{N+2}$ as
\vspace{-2mm}
\begin{equation}
  J(x) =
  \big(\lambda_{\pspec}(\tword(\traj)), \ftime(\traj)\big).
  \label{eq:cost-function}
  \vspace{-2mm}
\end{equation}
Recall from Section \ref{sec:prelim} that
$\ftime(\traj)$ denotes the final time of trajectory $\traj$.
As a result, the last coordinate of the cost function $J$ corresponds to
the minimum-time objective whereas
the first coordinate, $\lambda_{\pspec}(\tword(\traj))$, corresponds to
the level of unsafety of $\traj$ with respect to $\pspec$.

Using the cost function $J$, we formally define the minimum-violation planning as follows.
\begin{problem}[Minimum-Violation Planning]
  Based on the standard lexicographical order, compute an optimal trajectory
  $\traj^* = \arg\min_{\traj \in \trajs} J(\traj)$.
  \label{prob:orig}
\end{problem}

\begin{remark}
  We choose the minimum-time objective as indicated by the maneuver cost function $\ftime$
  for the simplicity of the presentation.
  Our approach also applies to other maneuver costs, including the control effort,
  with some minor modifications.
\end{remark}



\section{SOLUTION}
\label{sec:theoretical}

Reference \cite{Castro:2013:CDC} solves Problem \ref{prob:orig} by
constructing a weighted finite automaton $\automaton$
that is the product of weighted finite automata,
each corresponding to an atomic safety rule
$\spec \in \Phi$.
The weights on the transitions of $\automaton$ are defined such that
the weight of the shortest accepting run over any word $\tword$
is the level of unsafety of $\tword$.
The product $\kripke \otimes \automaton$
of the Kripke structure $\kripke$ and $\automaton$
is incrementally constructed.
It can be shown that Problem \ref{prob:orig} is equivalent to finding a shortest path
in $\kripke \otimes \automaton$.
%

As the size of $\automaton$ is exponential in the length of $\spec$ \cite{Baier:2008:Principles},
our approach avoids constructing the product
$\kripke \otimes \automaton$ to reduce computational complexity.
Instead, we translate an $\siFLTLX$ formula over $\props$
into an $\siFLTL$ formula over $\props \times \props$.
As will be discussed later, this translation allows us to incrementally construct and maintain only
the Kripke structure $\kripke$ (as opposed to $\kripke \otimes \automaton$ as in the aforementioned work),
and compute the weights of its transitions based on the satisfaction of propositional formulas
of the consecutive states and the time duration of the transitions.
As a result, it allows temporal logic specifications to be handled with
the same computational complexity as traditional motion planning algorithms
such as RRT* and RRG.

\begin{definition}[$\siFLTL$]
  An $\siFLTL$ formula over $\props \times \props$ is an $\siFLTLX$ formula
  $\spec = \always P$ where $P$ is a propositional logic formula over
  $\props \times \props$.
  \label{def:siFLTL}
\end{definition}

A propositional logic formula $P$ over
$\props \times \props$ is interpreted over a pair
$(l, l') \in 2^\props \times 2^\props$
with the satisfaction relation $\satisfies$ defined as follows.
For $p, p' \in \props \union \{\true, \false\}$ and
$(l, l') \in 2^\props \times 2^\props$,
$(l, l') \satisfies (p, p')$ if and only if
$l \satisfies p$ and $l' \satisfies p'$.
Here, for any $l \in 2^\props$, we have
$l \satisfies \true$,
$l \not\satisfies \false$,
and for any $p \in \props$,
$l \satisfies p$ if and only if $p \in l$.
The logic connectives are defined as in the standard propositional logic.

Based on the semantics of FLTL,
given a finite word $\word = l_0 l_1 \ldots l_n \in (2^\props)^{n+1}$
and an $\siFLTL$ formula $\spec = \always P$ over $\props \times \props$,
we say that $\word$ satisfies $\spec$,
written $\word \satisfies_{\props \times \props} \spec$
if and only if $(l_i, l_{i+1}) \satisfies P$ for all $i \in \naturalsle{n-1}$
and $(l_n, l_n) \satisfies P$.
Note that the terminal condition $(l_n, l_n) \satisfies P$ results
from the assumption that $\spec$ is stutter-invariant, which ensures that
$w \satisfies \spec$ if and only if
$w' = l_0 l_1 \ldots l_n l_n \satisfies \spec$.

The level of unsafety of a finite timed word
$\tword = l_{0}^{t}l_{1}^{t} \ldots l_{n}^{t}$
with respect to an $\siFLTL$ formula
$\spec$ over $\props \times \props$ is defined by
\begin{equation}
        \lambda(\tword, \spec) =
        \mkern-20mu
        \sum_{i \in \naturalsle{n} \hspace{1mm}|\hspace{1mm} (l_i, l_{i+1}) \not\satisfies P}
        \mkern-20mu
        \tilde{\lambda}(l_{i}^{t}, P),
        \label{eq:unsafety-individual-siFLTL}
\end{equation}
where
$l^t_i = (l_i, d_i) \in 2^\props \times \reals_{\geq 0}$
for all $i \in \naturalsle{n}$,
$l_{n+1}^{t} = (l_n,0)$,
$\tilde{\lambda}(l_{i}^{t}, P) = d_{i}$ if $(l_i, l') \not\satisfies P$ for all $l' \in 2^\props$;
otherwise $\tilde{\lambda}(l_{i}^{t}, P) = 1$.

\subsection{Conversion of $\siFLTLX$ to $\siFLTL$ }
%
Given an $\siFLTLX$ formula $\spec$ over $\props$,
we define an operation $\denext$ that
constructs an $\siFLTL$ formula over $\props \times \props$ from $\spec$ by
replacing each instance of $p$ in $\spec$ with $(p, \true)$
and replacing each instance of $\next p$ in $\spec$ with $(\true, p)$
for all $p \in \props$.
For example, consider an $\siFLTLX$ formula $\spec$ defined in (\ref{ex:spec}).
The corresponding $\siFLTL$ formula over $\props \times \props$ is given by

\vspace{-5mm}
\begin{equation}
  \denext(\spec) = \always
  \Big( (p, \true) \implies \big((\true, p) \oor (\true, p') \big) \Big).
  \label{ex:denext}
\end{equation}

\begin{lemma}
Let $\props = \{p_0, p_1, \ldots, p_n\}$
and $\props' = \{q_0, q_1, \ldots, q_{n'}\}$
be sets of propositions.
Let $P$ be a propositional formula over $\props \union \props'$
and $P'$ be a propositional formula over $\props \times \props'$
that is constructed from $P$
by replacing each instance of $p_i$ by $(p_i, \true)$
and replacing each instance of $q_j$ by $(\true, q_j)$
for all $i \in \naturalsle{n}$, $j \in \naturalsle{n'}$.
Then, for any $l \subseteq \props$ and $l' \subseteq \props'$,
$l \union l' \satisfies P$ if and only if
$(l, l') \satisfies P'$.
\label{lemma:propositional_formula}
\end{lemma}
\begin{proof}
Consider arbitrary $l \subseteq \props$ and $l' \subseteq \props'$,
$i \in \naturalsle{n}$ and $j \in \naturalsle{n'}$.
It follows directly from the definition of $\true$ that
\begin{inparaenum}[(a)]
\item $l \satisfies p_i$ if and only if
$(l, l') \satisfies (p_i, \true)$, and
\item, $l' \satisfies q_j$ if and only if
$(l, l') \satisfies (\true, q_j)$.
\end{inparaenum}
As a result, we can conclude from the construction of $P'$
and the sematics of propositional logic that
$l \union l' \satisfies P$ if and only if $(l, l') \satisfies P'$.
\end{proof}

\begin{lemma}
Let $\PX$ be defined as in Definition \ref{def:siFLTLX}.
Consider a propositional logic formula $P'$ over $\props \times \props$ that
is constructed from $\PX$ by
replacing each instance of $p$ in $\PX$ with $(p, \true)$ and
replacing each instance of $\next p$ in $\PX$ with $(\true, p)$
for all $p \in \props \cup \{\true, \false\}$.
For any arbitrary $l, l' \in 2^\props$, we have
$l l' \satisfies \PX$ if and only if
$(l, l') \satisfies P'$.
\label{lemma:next_formula}
\end{lemma}
\begin{proof}
Let $\props = \{p_0, p_1, \ldots, p_n\}$.
Define $\props' = \{q_0, q_1, \ldots, q_n\}$.
Let $P$ be a propositional formula over $\props \union \props'$
that is constructed from $\PX$ by
replacing each instance of $\next p_i$ by $q_i$ for all $i \in \naturalsle{n}$.

Consider arbitrary $l, l' \in 2^\props$.
Let $\props_{l'} = \big\{q_i \hspace{1mm}|\hspace{1mm} p_i \in l'\big\} \subseteq \props'$.
Based on the definition of the $\next$ operator, we can conclude that
$l l' \satisfies \PX$ if and only if
$l \union \props_{l'} \satisfies P$.

By construction, $P'$ is obtained from $P$
by replacing each instance of $p_i$ in $P$ by $(p_i, \true)$
and replacing each instance of $q_i$ in $P$ by $(\true, p_i)$
for all $i \in \naturalsle{n}$.
As a result, we can conclude using Lemma \ref{lemma:propositional_formula} and
the definition of $\props_{l'}$ that
$l \union \props_{l'} \satisfies P$ if and only if
$(l, l') \satisfies P'$.
Combining this with the result from the previous paragraph, we obtain
$l l' \satisfies \PX$ if and only if
$(l, l') \satisfies P'$.
\end{proof}

We now establish the equivalence of
the level of unsafety with respect to an $\siFLTLX$ formula over $\props$ and
the level of unsafety with respect to the corresponding $\siFLTL$ formula over $\props \times \props$.

\begin{lemma}
  For any finite timed word $\tword$
  and any $\siFLTLX$ formula $\spec$ over $\props$,
  \begin{equation}
    \lambda(\tword, \spec) = \lambda(\tword, \denext(\spec)).
  \end{equation}
  \label{lemma:unsafety-individual}
\end{lemma}
\vspace{-3mm}
\begin{proof}
  This result can be trivially derived from Lemma \ref{lemma:next_formula} and the definitions of
  the level of unsafety (\ref{eq:unsafety-individual-siFLTLX}) and (\ref{eq:unsafety-individual-siFLTL}).
\end{proof}

Finally, we construct the prioritized safety specification
$\hat{\pspec} = (\props \times \props, \hat{\Phi}, \hat{\Psi}, \hat{\priority})$
with each atomic safety rule obtained from that of $\pspec$
by applying $\denext$ operation.
Formally,
$\hat{\Phi} = \big\{ \denext(\spec) \hspace{1mm}|\hspace{1mm} \spec \in \Phi \big\}$,
$\hat{\Psi} = (\hat{\Psi}_0, , \hat{\Psi}_1, \ldots, \hat{\Psi}_N)$,
$\hat{\Psi}_i = \big\{ \denext(\spec) \hspace{1mm}|\hspace{1mm} \spec \in \Psi_i \big\}$ for all $i \in \naturalsle{N}$, and
$\priority(\spec) = \hat{\priority}(\denext(\spec))$ for all $\spec \in \Phi$.
The level of unsafety of a finite timed word $\tword$ with respect to $\hat{\pspec}$ is defined
following (\ref{eq:unsafety-prioritized}), (\ref{eq:unsafety-set}) as
$\lambda_{\hat{\pspec}}(\tword) = (\lambda_{\hat{\pspec}}(\tword, \hat{\Psi}_0), \ldots, \lambda_{\hat{\pspec}}(\tword, \hat{\Psi}_N)) \in \reals^{N+1}$,
where for each $i \in \naturalsle{N}$,
$\lambda_{\hat{\pspec}}(\tword, \hat{\Psi}_i) = \sum_{\spec \in \hat{\Psi}_i} \hat{\priority}(\spec) \lambda(\tword, \spec)$.
Based on the construction of $\hat{\pspec}$ and Lemma \ref{lemma:unsafety-individual},
we obtain the following result,
which allows us to replace $\pspec$ with $\hat{\pspec}$.

\begin{proposition}
  For any finite timed word $\tword$, $\lambda_{\pspec}(\tword) = \lambda_{\hat{\pspec}}(\tword)$.
  \label{prop:siFLTLX-siFLTL-equivalence}
\end{proposition}


\subsection{Incremental Construction of Weighted Kripke Structure}
We follow a sampling-based procedure described in \cite{Karaman:2011:IJRR}
to incrementally construct a Kripke structure $\kripke$
as a finite state representation of the dynamical system (\ref{eq:dynamical_system}).
The main difference is that we augment $\kripke$ with weights on its transitions.
The weights are picked such that the sum of the weights on any finite trace
$\tau$ of $\kripke$ is the level of unsafety of the finite timed word generated by
the trajectory of (\ref{eq:dynamical_system}) corresponding to $\tau$.
As opposed to \cite{Castro:2013:CDC}, we do not construct the weighted product automaton
$\kripke \otimes \automaton$ where $\automaton$ is created by combining all the automata,
each corresponding to each $\spec \in \Phi$.

\begin{definition}[Weighted Kripke Structure]
  A \emph{weighted Kripke structure} is a tuple
  \begin{equation}
    \wkripke = (\kstates, \ksinit, \krel, \kprops, \klabeling, \kweight),
    \label{eq:weighted-kripke}
  \end{equation}
  where
  $\kstates$, $\ksinit$, $\krel$, $\kprops$ and $\klabeling$ are defined as in
  Definition \ref{def:kripke} and
  $\kweight : \krel \to \reals_{\geq 0}^n$ for some $n \in \naturals$
  is a function assigning a transition cost to each transition in $\krel$.
  \label{def:weighted-kripke}
\end{definition}

A \emph{finite trace} of $\wkripke$ is defined as that of $\kripke$.
Given a finite trace $\tau = s_0 s_1 \ldots s_n$,
we define the weight of $\tau$ as
$\mathcal{W}(\tau) = \sum_{i \in \naturalsle{n-1}} \kweight(s_i, s_{i+1})$.
%
For any $s \in \kstates$, define
$\Traces(\wkripke, s) = \big\{
\tau = s_0 s_1 \ldots s_n \hspace{2mm}|\hspace{2mm}
n \in \naturals,
s_n = s,
\tau \text{ is a finite trace of } \wkripke
\big\}$
to be the set of all the finite traces of $\wkripke$ that end at $s$.

Various sampling-based algorithms such as RRT*, RRG,
and their $k$-nearest variants
can be employed to incrementally construct $\wkripke$
\cite{Karaman:2011:IJRR}.
The key difference between these algorithms
lies in the connections of states.
In particular, the RRT* algorithm maintains a tree structure rather
than a graph as in RRG,
ensuring that each state only has at most one parent.
It maintains an upper bound on the cost $\costtocome(s)$ of the unique path
from the initial state to each state $s \in \kstates$.
Algorithm \ref{alg:kripke-construction} provides a common
template for incrementally constructing $\wkripke$,
based on the following primitive procedures.

\begin{enumerate}[a)]
\item \emph{Sample:}
  $\sample : \naturals \to \states$ is a function that generates
  independent, identically distributed samples
  from a uniform distribution supported over $\states$.
\item \emph{Add:}
  Given a state $s$, $\addstate(s)$ adds $s$ to $\kstates$, i.e.,
  it executes $\kstates \leftarrow \kstates \cup \{s\}$.
  For RRT*, it also sets $\costtocome(s) = 0$ if $s = \sinit$;
  otherwise $\costtocome(s) = \infty$.
\item \emph{Steer:}
  Given states $s, s' \in \states$, $\steer(s, s')$ returns the set of trajectories
  $\traj : [0, T] \to \states$ of (\ref{eq:dynamical_system}) such that
  $\traj(0) = s$,
  $\traj(T) = s'$, and
  $\mathcal{T}(\traj)$ exists and is finite.
\item \emph{Nearest neighbors:}
  $\near : \states \to 2^{\kstates}$ computes the set of nearest neighbors.
  When applying the RRT* or RRG algorithm, we let
  \vspace{-2mm}
  \begin{equation*}
    \near(s) =
    \left\{
      s' \in \kstates \hspace{2mm}\Big|\hspace{2mm}
      \|s' - s\|_2 \leq \left(\gamma \frac{\log m}{m}\right)^{1/D}
    \right\},
  \end{equation*}
  where $m$ is the cardinality of $\kstates$,
  and $\gamma$ and $D$ are constants that depend on the dimension $d$ of the state space
  and the Lebesgue measure of $\states$.
  For the case of $k$-nearest RRG or $k$-nearest RRT*,
  $\near(s)$ returns $k$ nearest neighbors of $s \in \states$
  where $k > \gamma' \log(n)$ for some constant $\gamma'$.
  We refer the reader to \cite{Karaman:2011:IJRR,Solovey:2020:ICRA} for the definitions of $\gamma$, $\gamma'$, and $D$.
\item \emph{Transition cost:}
  For any trajectory $\traj : [0, T] \to \states$ of (\ref{eq:dynamical_system}),
  we define the cost of $\traj$ as
  $\mathcal{C}(\traj) = \big(\lambda_{\hat{\pspec}}(\tword(\traj)), T\big) \in \reals^{N+2}$.
  Note that according to (\ref{eq:cost-function}) and Proposition \ref{prop:siFLTLX-siFLTL-equivalence},
  $\mathcal{C}(\cdot)$ corresponds to the cost function defined in Problem \ref{prob:orig}.
  Let $\tword(x) = (l_0, d_0)(l_1, d_1) \ldots (l_n, d_n)$.
  Thanks to (\ref{eq:unsafety-individual-siFLTL}),
  $\lambda_{\hat{\pspec}}(\tword(\traj))$ can be computed by simply evaluating whether
  $(l_i, l_{i+1})$
  satisfies the propositional formula corresponding to each
  atomic safety rule for each $i \in \naturalsle{n}$, with $l_{n+1} = l_n$.
  For any states $s, s' \in \states$,
  the transition cost from $s$ to $s'$ is then defined for the case where
  $\steer(s, s') \not= \emptyset$ as
  $\tcost(s, s') = \min_{\traj \in \steer(s, s')} \mathcal{C}(\traj)$.
\item \emph{Connect:}
  Given states $s, s' \subseteq \kstates$,
  $\connect(s, s')$ updates the relevant elements
  based on a transition from $s$ to $s'$.
  For RRG, $\connect(s, s')$ simply adds the transition
  $(s, s')$ to $\krel$ and set $\kweight(s, s') = \tcost(s, s')$
  as shown in Algorithm~\ref{alg:connect-RRG}.
  In contrast, as shown in Algorithm~\ref{alg:connect-RRT},
  the transition $(s, s')$ is added for RRT*
  only if it improves the cost to reach $s'$ from $\sinit$
  (Line \ref{lst:rrt-update-condition}).
  If so, existing transitions to $s'$ are removed
  (Line \ref{lst:rrt-transition})
  %
  %
  and $\costtocome(s')$ is updated (Line \ref{lst:rrt-update-cost-start}).
  By propagating the change in $\costtocome(s')$ down the tree structure,
  we obtain the RRT\# algorithm, which ensures that the \textit{promising} vertices
  (i.e., those that have the potential to be part of the optimal solution) are consistent,
  i.e., $\costtocome(\tilde{s})$ is the cost of the unique path from the initial state
  to a promising vertex $\tilde{s} \in \kstates$.
  This approach has been shown to improve the convergence rate of RRT*.
  See \cite{Arslan:2013:Use} for more details on RRT\#.
  %
\end{enumerate}

Algorithm \ref{alg:kripke-construction} returns
the weighted Kripke structure $\wkripke_{n}$
after $n$ iterations as well as the set
$\kSgoal = \Sgoal \intersect \kstates$ of the sampled goal states.
For each state $s \in \kstates$,
let $\tau(\wkripke_{n}, s) \in \Traces(\wkripke_{n}, s)$
denote an optimal trace of $\wkripke_{n}$ that ends at $s$, i.e.,
$\mathcal{W}(\tau(\wkripke_{n}, s)) \leq \mathcal{W}(\tilde{\tau})$
for all $\tilde{\tau} \in \Traces(\wkripke_{n}, s)$.
We define $\ksgoal = \arg\min_{s \in \kSgoal}\mathcal{W}(\tau(\wkripke_{n}, s))$.
Note that for the case of RRT\#,
$\ksgoal = \arg\min_{s \in \kSgoal}\costtocome(s)$ and
for any promising vertex $s \in \kstates$,
$\mathcal{W}(\tau(\wkripke_{n}, s)) = \costtocome(s)$ and
$\tau(\wkripke_{n}, s)$ can be obtained
by following the unique parent of each state backward,
starting from $s$ to the initial state, i.e.,
$\tau(\wkripke_{n}, s) = s_0 s_1 \ldots s_m$ for some $m \in \naturals$ such that
$s_0 = \sinit$,
$s_m = s$, and
$s_i = \parent(s_{i+1})$ for all $i \in \naturalsle{m-1}$.
Here, $\parent(\tilde{s})$ is a unique state with
$(\parent(\tilde{s}), \tilde{s}) \in \krel$.
When the connections are constructed based on the RRT* or RRG algorithm,
$\ksgoal$ and $\tau(\wkripke_{n}, s)$ can be obtained using, e.g.,
the Dijkstra's shortest path algorithm.

\begin{algorithm}[h]
  \SetInd{0.5em}{0.5em}
  $\kstates \leftarrow \emptyset$; $\krel \leftarrow \emptyset$; $\kSgoal \leftarrow \emptyset$\;
  $\addstate(\sinit)$\;
  \ForEach{$i \in \naturalsle{n}$}{ \label{lst:planning-loop-start}
    $\snew \leftarrow \sample(i)$\;
    $S_{\mathsf{near}} = \near(\snew)$\;
    $\addstate(\snew)$\;
    \ForEach{$s \in S_{\mathsf{near}}$}{
      \If{$\steer(s, \snew) \not= \emptyset$}{
        $\connect(s, \snew)$\;
      }
    }
    \ForEach{$s \in S_{\mathsf{near}}$}{ \label{lst:rewiring-start}
      \If{$\steer(\snew, s) \not= \emptyset$}{
        $\connect(\snew, s)$\;
      }
    }
    \label{lst:rewiring-finish}
    \If{$\snew \in \Sgoal$}{
      $\kSgoal = \kSgoal \union \{\snew\}$\;
    } \label{lst:planning-loop-finish}
  }
  \Return{$\wkripke_n = (\kstates, \sinit, \krel, \props, \labeling, \kweight)$, $\kSgoal$}
  \caption{Minimum-violation planning.}
  \label{alg:kripke-construction}
\end{algorithm}
\vspace{-5mm}
\begin{algorithm}[h]
  \SetInd{0.5em}{0.5em}
  $\krel \leftarrow \krel \cup \{(s, s')\}$\;
  $\kweight(s, s') = \tcost(s, s')$
  \caption{$\connect(s, s')$ for RRG.}
  \label{alg:connect-RRG}
\end{algorithm}
\vspace{-5mm}
\begin{algorithm}[h]
  \SetInd{0.5em}{0.5em}
  \If {$\costtocome(s) + \tcost(s,s') < \costtocome(s')$} {\label{lst:rrt-update-condition}
    $\krel \leftarrow \big(\krel \setminus \{(s_{1}, s_{2}) \in \krel
    \ | \ s_{2} = s'\}\big) \cup \{(s, s')\}$\;\label{lst:rrt-transition}
    $\kweight(s, s') = \tcost(s, s')$\;
    $\costtocome(s') = \costtocome(s) + \tcost(s, s')$\;\label{lst:rrt-update-cost-start}
  }
  \caption{$\connect(s, s')$ for RRT*.}
  \label{alg:connect-RRT}
\end{algorithm}

Let $c^* = \min_{s \in \Sgoal} \tcost(\sinit, s)$ be the cost of an optimal trajectory of (\ref{eq:dynamical_system})
from $\sinit$ to $\Sgoal$ based on the cost function $\mathcal{C}$.
The following result can be directly derived from the asymptotic optimality of the
RRG and RRT* algorithms \cite{Karaman:2011:IJRR}
and the boundedness of $\tcost$ for the case where
$\PX$ in (\ref{eq:siFLTLX}) is a propositional formula for all $\spec \in \Phi$.
\begin{lemma}
  If $\PX$ in (\ref{eq:siFLTLX}) is a propositional formula for all $\spec \in \Phi$,
  then $\mathcal{W}(\tau(\wkripke_{n}, \ksgoal))$ converges to $c^*$ almost surely,
  i.e.,
  \begin{equation*}
    \mathbb{P}\left( \left\{ \lim_{n \to \infty} \mathcal{W}(\tau(\wkripke_{n}, \ksgoal))
        = c^* \right\} \right) = 1.
  \end{equation*}
  \label{lemma:asymtotic-optimality-cost}
\end{lemma}

We now establish the equivalence of the transition cost $\tcost$ and the original cost function $J$ of Problem (\ref{prob:orig}).
\begin{lemma}
  $c^* = \min_{\traj \in \trajs} J(\traj)$.
  \label{lemma:J-cost-equivalence}
\end{lemma}
\begin{proof}
  This result follows directly from Proposition \ref{prop:siFLTLX-siFLTL-equivalence} and the definitions of $J$ and $\tcost$.
\end{proof}

Consider an arbitrary $s \in \kstates$.
Let $\tau(\wkripke_{n}, s) = s_0 s_1 \ldots s_m$ for some $m \in \naturals$.
For each $i \in \naturalsle{m-1}$, let $\traj_{s,i} = \arg\min_{\traj \in \steer(s_i, s_{i+1})} \mathcal{C}(\traj)$, i.e.,
$\traj_{s,i}$ is a trajectory of (\ref{eq:dynamical_system}) from $s_i$ to $s_{i+1}$ with $\mathcal{C}(\traj_{s,i}) = \tcost(s_i, s_{i+1})$.
A trajectory $\traj_s$ of (\ref{eq:dynamical_system}) from $\sinit$ to $s$ can be constructed from $\tau(\wkripke_{n}, s)$ by concatinating $\traj_{s,i}$.
Formally, $\traj_s : [0, \sum_{i \in \naturalsle{m-1}} T_i] \to \states$ such that
$\traj_s(t + \sum_{k = 0}^{i-1} T_k) = \traj_{s,i}(t)$ for all $t \in [0, T_i]$ and $i \in \naturalsle{m-1}$
where $T_i = \ftime(\traj_{s,i})$.

\begin{lemma}
  For any $s \in \kstates$,
  $\mathcal{W}(\tau(\wkripke_{n}, s)) = J(\traj_s)$.
  \label{lemma:JK-J-equivalence}
\end{lemma}
\begin{proof}
  Let $\tau(\wkripke_{n}, s) = s_0 s_1 \ldots s_m$.
  By construction, $\mathcal{W}(\tau(\wkripke_{n}, s)) =
  \sum_{i \in \naturalsle{m-1}} \tcost(s_i, s_{i+1})$.
  Additionally, from the definition of $\mathcal{C}$, $J$
  and Proposition \ref{prop:siFLTLX-siFLTL-equivalence},
  $J(\traj_s) = \sum_{i \in \naturalsle{m-1}} \mathcal{C}(\traj_{s,i})$.
  We can then conclude from the definition of $\traj_{s,i}$ that
  $J(\traj_s) = \sum_{i \in \naturalsle{m-1}} \tcost(s_i, s_{i+1})$.
\end{proof}

Combining Lemma \ref{lemma:asymtotic-optimality-cost}--\ref{lemma:JK-J-equivalence},
we obtain the asymtotic optimality of Algorithm \ref{alg:kripke-construction}.

\begin{proposition}
  Let $\traj^*$ be a solution of Problem \ref{prob:orig} and $J^* = J(\traj^*)$.
  If $\PX$ in (\ref{eq:siFLTLX}) is a propositional formula for all $\spec \in \Phi$,
  then $J(\traj_{\ksgoal})$ converges to $J^*$ almost surely.
\end{proposition}

If there exists $\spec \in \Phi$ such that $\PX$ is not a propositional formula,
then the asymptotic optimality of Algorithm~\ref{alg:kripke-construction} cannot be guaranteed
as $\tcost$ is not necessarily bounded.
However, Lemma~\ref{lemma:JK-J-equivalence} ensures that
$\tau(\wkripke_{n}, \ksgoal)$ is an optimal trajectory
among those in $\wkripke_{n}$.

The analysis in \cite{Castro:2013:CDC} shows that the computational complexity
of the original Minimum-Violation RRT* algorithm is $\mathcal{O}(K^2 n \log n)$ where
$K$ is the number of states in the weighted finite automaton $\automaton$,
which is exponential in the length of the specification.
As Algorithm \ref{alg:kripke-construction} is a special case of the RRT* and RRG algorithms
where the transition cost corresponds to the level of unsafety of the transition,
it shares the same computational complexity of $\mathcal{O}(n \log n)$ as that of
the RRT* and RRG algorithms,
which is the same as that of the original Minimum-Violation RRT* algorithm with $K = 1$.
As in\cite{Castro:2013:CDC}, this analysis relies on the assumption
that the complexity of $\connect(s, s')$ is $\mathcal{O}(1)$ for all $s, s' \in \kstates$.



\section{PRACTICAL CONSIDERATIONS}
\label{sec:practical}

The asymptotic optimality of Algorithm \ref{alg:kripke-construction}
is essential in many safety-critical applications as it ensures that
a sufficiently safe trajectory will be found,
if one exists,
given sufficient computation time.
For autonomous vehicles, however, the available computation time is often limited
due to the dynamic nature of the environments in which they operate.
Such environments include not only relatively static features such as
road markings, constructions, weather conditions, etc.,
but also dynamic features arising from other agents (vehicles, pedestrians, animals, etc.)
sharing the road.
An optimal trajectory with respect to the environment at time $t \in \reals_{\geq 0}$ may become
the least safe option with respect to the environment at time $t + \epsilon$,
even for small $\epsilon \in \reals_{\geq 0}$,
especially when there is a drastic change in the environment.
Examples of such situations include
\begin{inparaenum}[(a)]
\item a newly detected object shows up,
\item another agent violates the right of way, and
\item a vehicle that is initially parked starts to move off
  while the autonomous vehicle is overtaking it.
\end{inparaenum}
In fact, a similar situation to the latter led to an accident between the Cornell and the MIT autonomous vehicles
during the 2007 DARPA Urban Challenge \cite{Fletcher:2008:JFR}.

Reactive synthesis \cite{Wongpiromsarn:2012:TAC,Kress-Gazit:2008:CASE,Kress-Gazit:2011:RAM} and
probabilistic synthesis \cite{Wongpiromsarn:2012:CDC,Ulusoy:2014:IJRR,Lahijanian:2012:TR}
have been applied to handle dynamic environments.
Roughly, in reactive synthesis, a control policy is constructed to ensure that
the system satisfies its specification for all valid environment behaviors.
In contrast, probabilistic synthesis considers a probabilistic model of the environment
and constructs a control policy that maximizes the probability
that the system satisfies its specification.
Both approaches assume a good understanding of the environment:
Reactive synthesis requires the knowledge of all possible behaviors of the environment,
whereas probabilistic synthesis relies on an accurate probabilistic model of the environment.
The control protocol synthesized by these approaches could be invalid
in that the system could be left with no valid trajectory if such assumptions
do not hold.

A key advantage of minimum-violation planning compared to these approaches
is in handling unexpected or unmodeled environment behaviors:
As long as there exist trajectories from $\sinit$ to a state in $\Sgoal$
in $\wkripke_{n}$, the algorithm always returns the safest one.
Hence, the richness of $\wkripke_{n}$ is a crucial factor for
successful applications of this approach.

As autonomous vehicles are required to respond quickly to changes in the environment,
$\wkripke_{n}$ needs to be updated in each planning iteration.
To ensure the richness of $\wkripke_{n}$,
we maintain $\wkripke_{n}$ from the previous planning iteration
instead of rebuilding it from scratch.
Then, each planning iteration updates $\sinit$ and $\kweight$
based on the current state of the vehicle and
the most recently observed environment.
Additionally, due to changes in $\kweight$,
RRG-based algorithms are potentially more preferable
than RRT*-based algorithms.

In summary, we initialize $\wkripke_{n}$ with $\kstates = \{\sinit\}$
and $\krel = \emptyset$.
Then, each planning iteration performs the following procedures.
\begin{enumerate}[(i)]
\item Update $\sinit$ and $\kweight$ based on the current state of the system.
\item Augment $\wkripke_{n}$ based on lines
  \ref{lst:planning-loop-start}--\ref{lst:planning-loop-finish}
  of Algorithm \ref{alg:kripke-construction}.
\item Extract an optimal trajectory in $\wkripke_{n}$ using a graph
  search algorithm (e.g., Dijkstra or A*).
\end{enumerate}

Finally, Algorithm \ref{alg:kripke-construction} can be adapted to other defitions
of the level of unsafety by modifying the transition cost function $\mathcal{C}$
such that $J(\traj_s) = \sum_{i \in \naturalsle{m-1}} \mathcal{C}(\traj_{s,i})$
for all $s \in \kstates$ where $s_0 s_1 \ldots s_m = \tau(\wkripke_{n}, s)$.
In this case, it is easy to show that Lemma \ref{lemma:J-cost-equivalence}
and Lemma \ref{lemma:JK-J-equivalence} still hold;
thus, an optimal trace of $\wkripke_{n}$ to a goal state as extracted by a graph search algorithm
still corresponds to a safest trajectory among all the options in $\wkripke_{n}$.
In fact, we have experimented with other definitions of
the level of unsafety as suggested by the rulebooks framework \cite{Censi:2019:ICRA},
including non-additive costs \cite{Slutsky:2020:WAFR}.
Certain definitions, however, invalidate the assumption of
Lemma \ref{lemma:asymtotic-optimality-cost}
and as a result, lead to the loss of asymptotic optimality guarantee
similar to the case where $\PX$ in (\ref{eq:siFLTLX}) is not a propositional formula.
Refining the definition of the level of unsafety
as well as ensuring the richness of $\wkripke_{n}$
are subject to the current research.


\section{EXPERIMENTAL RESULTS}
\label{sec:results}

We consider an autonomous vehicle modeled by a Dubins car \cite{Lavalle:2006:Planning}:
$\dot{x} = \cos(\theta)$, $\dot{y} = \sin(\theta)$ and $\dot{\theta} = u$
where $u \in [-1, 1]$, $(x,y)$ is the position of the center of the rear axle
and $\theta$ is the heading of the vehicle.
The autonomous vehicle encounters a stationary vehicle while it is navigating a two-lane road
with a solid yellow center line as show in Figure~\ref{fig:exp-setup}.
The set of goal states is given by $\Sgoal = \{(x,y,\theta) \ | \ x \geq 37\}$.

\subsection{Prioritized Safety Specification}
For any $x, y, \theta \in \reals$, let
$\footprint(x,y,\theta) \subset \reals^{2}$ be the footprint of the autonomous vehicle
when the center of its rear axle is at $(x,y)$ and its heading is $\theta$.
Consider atomic propositions $\collision$, $\close$, $\onroad$ and $\inlane$,
representing the autonomous vehicle colliding with the stationary vehicle,
overlapping with the clearance zone, being fully on the road
and being fully within a correct lane, respectively.
Formally, the labeling function $\labeling$ is defined such that for any
$x, y, \theta \in \reals$,
\begin{inparaenum}[(a)]
\item $\collision \in \labeling(x, y, \theta)$ iff
  $\footprint(x, y, \theta) \intersect \obs \not= \emptyset$ where
  $\obs \subset \reals^{2}$ is the footprint of the stationary vehicle,
\item $\close \in \labeling(x, y, \theta)$ iff
  $\footprint(x, y, \theta) \intersect \clearance \not= \emptyset$ where
  $\clearance \subset \reals^{2}$ is the clearance zone around the stationary vehicle,
  constructed from $\obs$ and the required lateral and longitudinal clearance
  (see Figure~\ref{fig:exp-setup}),
\item $\onroad \in \labeling(x, y, \theta)$ iff
  $\footprint(x, y, \theta) \subseteq \road$ where
  $\road$ is the road, i.e., the area where a vehicle is allowed to drive, and
\item $\inlane \in \labeling(x, y, \theta)$ iff
  $\footprint(x, y, \theta) \subseteq \lane$ where
  $\lane$ is the right lane, i.e., the lane with the correct travel direction
  for the autonomous vehicle.
\end{inparaenum}

We consider the following atomic safety rules, each of which
can be expressed by an $\siFLTLX$ formula.
\begin{enumerate}[(i)]
\item \textbf{No collision:} $\spec_{1} = \always \neg \collision$.
\item \textbf{Staying on road:} $\spec_{2} = \always \onroad$.
\item \textbf{Obstacle clearance:} $\spec_{3} = \always \neg\close$.
\item \textbf{Lane keeping:} $\spec_{4} = \always \inlane$.
\end{enumerate}

The prioritized safety specification $\pspec = (\Pi, \Phi, \Psi, \priority)$
is defined as
$\props = \{\collision, \close, \onroad, \inlane\}$,
$\Phi = \{\spec_{1}, \ldots, \spec_{4}\}$,
$\Psi = \{ \{\spec_{1}\}, \{\spec_{2}\}, \{\spec_{3}, \spec_{4}\} \}$, and
$\priority(\spec_{i}) = 1$, for all $i$.

\subsection{Simulation Results}
\begin{figure}[htb]
  \centering
  \includegraphics[width=0.238\textwidth,trim={2.0cm 1.4cm 1.5cm 1cm},clip]{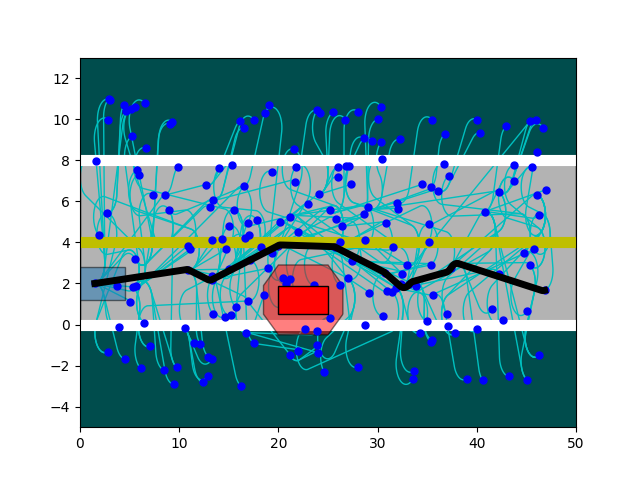}
  \includegraphics[width=0.238\textwidth,trim={2.0cm 1.4cm 1.5cm 1cm},clip]{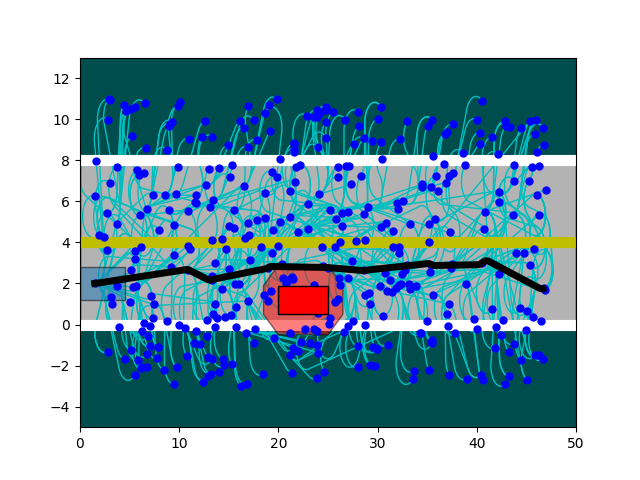}
  \includegraphics[width=0.238\textwidth,trim={2.0cm 1.4cm 1.5cm 1cm},clip]{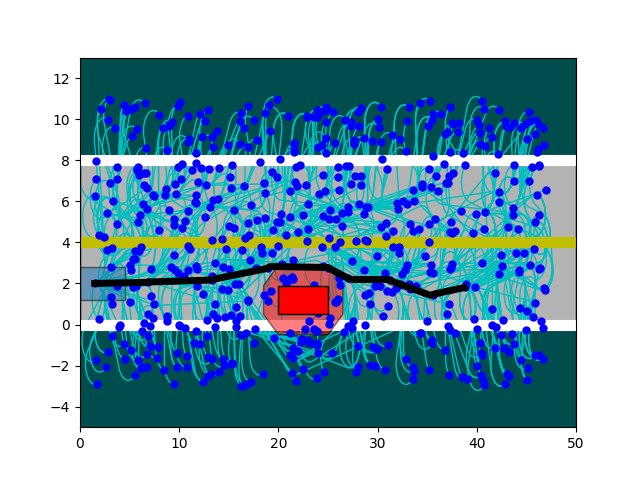}
  \includegraphics[width=0.238\textwidth,trim={2.0cm 1.4cm 1.5cm 1cm},clip]{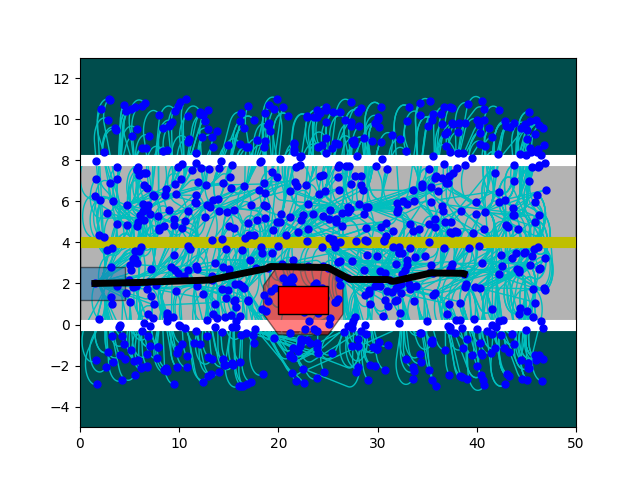}
  \vspace{-3mm}
  \caption{The states (dark blue dots) and their connections (light blue curves) in $\wkripke_{n}$
    and the optimal path (black curve) extracted at the end of
    the 10th (top left), 20th (top right), 30th (bottom left), and 40th (bottom right) iterations
    when Algorithm~\ref{alg:kripke-construction} is applied with RRT* connections.}
  \label{fig:exp-rrt-path}
  \vspace{-3mm}
\end{figure}

\begin{figure}[htb]
  \centering
  \includegraphics[width=0.34\textwidth,trim={0.5cm 0cm 1cm 1cm},clip]{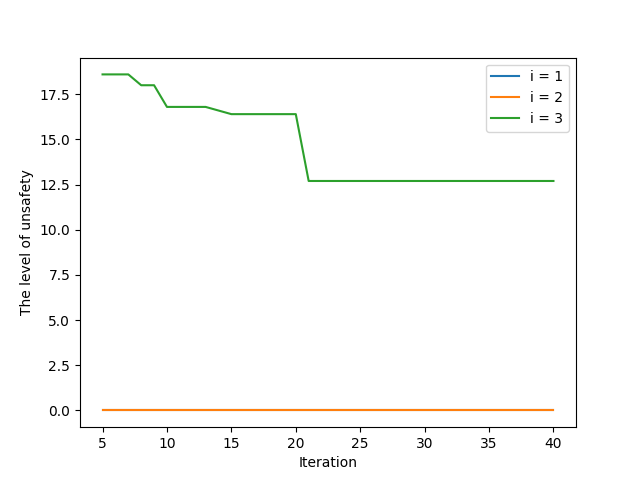}
  \includegraphics[width=0.34\textwidth,trim={0.5cm 0cm 1cm 1cm},clip]{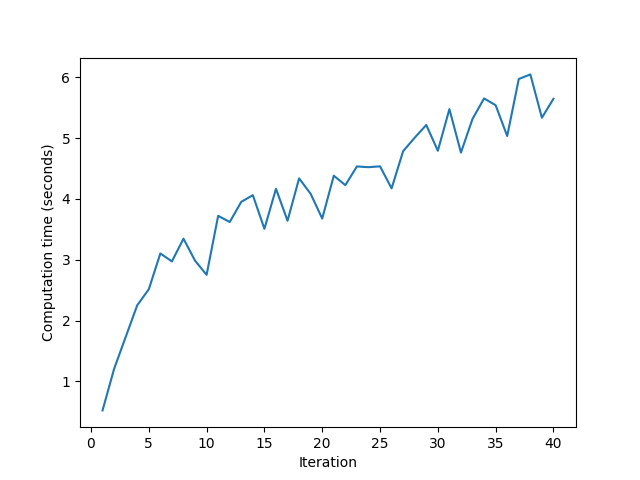}
  \vspace{-3mm}
  \caption{The level of unsafety of the optimal path with respect to
    $\Psi_{i}, i \in \{1, 2, 3\}$ and the computation time (seconds) of each iteration
    when Algorithm~\ref{alg:kripke-construction} is applied with RRT* connections.}
  \label{fig:exp-rrt-comp}
  \vspace{-1mm}
\end{figure}

Algorithm~\ref{alg:kripke-construction} was implemented in
TuLiP, a Python-based software toolbox \cite{Wongpiromsarn:2011:TuLiP}
and run on a laptop with Intel Core i7-10710U processor.
40 iterations of Line \ref{lst:planning-loop-start}--\ref{lst:planning-loop-finish}
of Algorithm~\ref{alg:kripke-construction} were run
with $n = 20$, i.e., 20 states were added in each iteration.
Figure~\ref{fig:exp-rrt-path} shows the
optimal traces $\tau(\wkripke_{n}, \ksgoal)$
when the connection is based on the RRT* algorithm.
The optimal cost $\mathcal{W}(\tau(\wkripke_{n}, \ksgoal))$ at the end of the 40th
iteration is $(0, 0, 12.7, 37.0)$.
The level of unsafety and computation time of each iteration is shown in
Figure~\ref{fig:exp-rrt-comp}.
%


\begin{figure}[htb]
  \centering
  \includegraphics[width=0.238\textwidth,trim={2.0cm 1.4cm 1.5cm 1cm},clip]{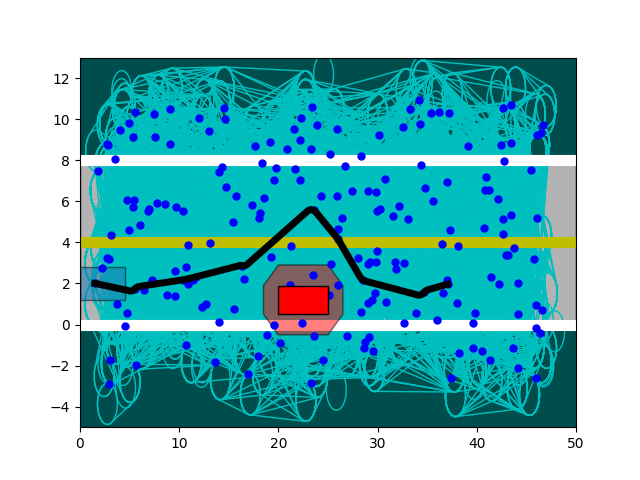}
  \includegraphics[width=0.238\textwidth,trim={2.0cm 1.4cm 1.5cm 1cm},clip]{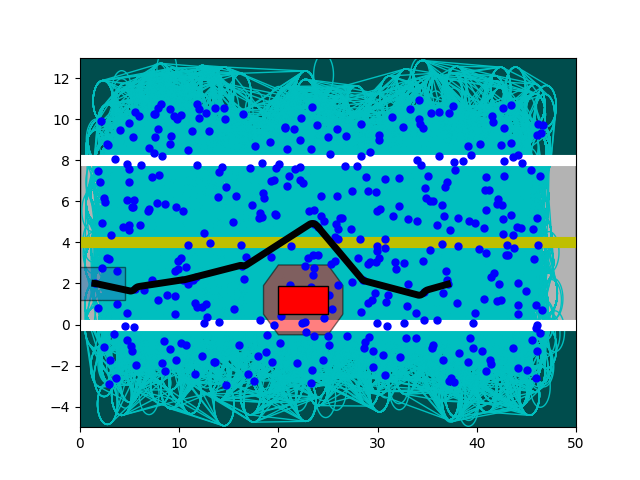}
  \includegraphics[width=0.238\textwidth,trim={2.0cm 1.4cm 1.5cm 1cm},clip]{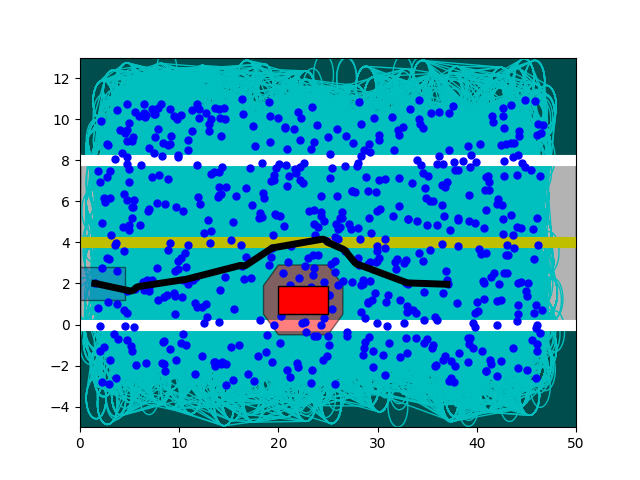}
  \includegraphics[width=0.238\textwidth,trim={2.0cm 1.4cm 1.5cm 1cm},clip]{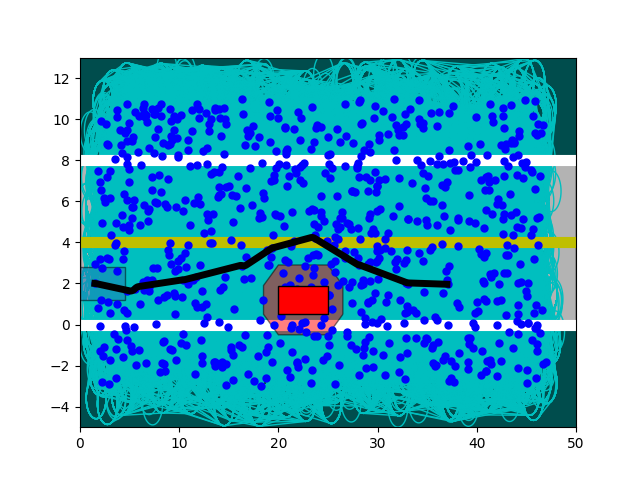}
  \vspace{-3mm}
  \caption{The states (dark blue dots) and their connections (light blue curves) in $\wkripke_{n}$
    and the optimal path (black curve) extracted at the end of
    the 10th (top left), 20th (top right), 30th (bottom left), and 40th (bottom right) iterations
    when Algorithm~\ref{alg:kripke-construction} is applied with RRG connections.}
  \vspace{-3mm}
  \label{fig:exp-rrg-path}
\end{figure}

\begin{figure}[htb]
  \centering
  \includegraphics[width=0.34\textwidth,trim={0.5cm 0cm 1cm 1cm},clip]{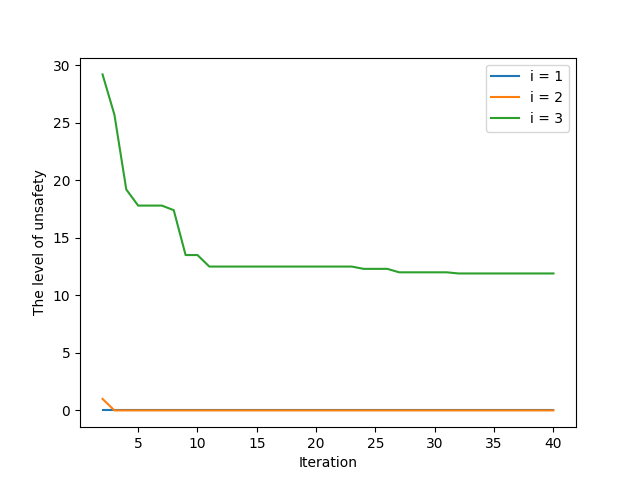}
  \includegraphics[width=0.34\textwidth,trim={0.5cm 0cm 1cm 1cm},clip]{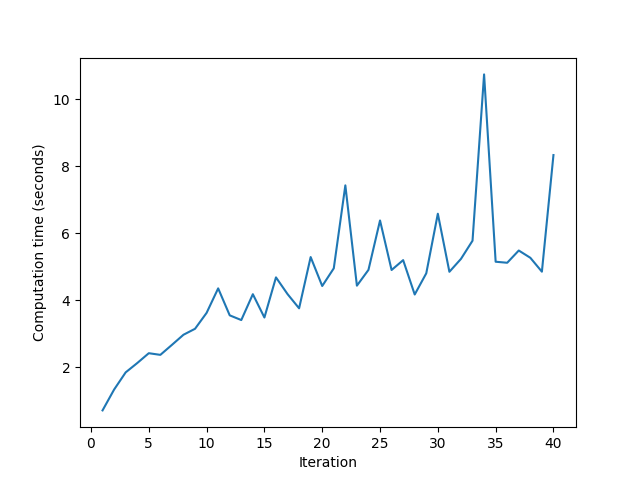}
  \vspace{-3mm}
  \caption{The level of unsafety of the optimal path with respect to
    $\Psi_{i}, i \in \{1, 2, 3\}$ and computation time (seconds) of each iteration
    when Algorithm~\ref{alg:kripke-construction} is applied with RRG connections.}
  \vspace{-3mm}
  \label{fig:exp-rrg-comp}
\end{figure}

The results when applying RRG connections are shown in Figure~\ref{fig:exp-rrg-path} and
Figure~\ref{fig:exp-rrg-comp}.
The optimal cost $\mathcal{W}(\tau(\wkripke_{n}, \ksgoal))$ at the end of the 40th
iteration is $(0, 0, 11.9, 35.5)$.
The Kripke structure constructed based on the RRG algorithm
includes significantly more connections than that constructed based on the RRT* algorithm.
For both cases, the level of unsafety with respect to $\Psi_{1}$ and $\Psi_{2}$
quickly converges to 0.


Note that the majority of the computation time is spent on computing labels along a trajectory.
This is similar to the case of traditional motion planning, where collision checking is typically
the main bottleneck \cite{Lavalle:2006:Planning}.
In fact, computing the violation of $\varphi_{1}$ is exactly the collision checking problem.
The specific implementation in this example performs expensive polygon operations to compute
the labels.
The computation time can be significantly reduced by employing more efficient polygon operations
and parallel computation.


\section{CONCLUSIONS}
\vspace{-1mm}
This paper introduced a class of LTL formulas that are
sufficiently expressive to describe traffic rules
such as lane-keeping, obstacle avoidance, etc.
Given traffic rules specified by these formulas
and their relative importance,
we proposed an incremental algorithm to compute a trajectory
for an autonomous vehicle to reach a given goal while
minimizing the level of unsafety with respect to the given rules.
Both the theoretical guarantees and practical considerations
were discussed.
Simulation results for the vehicle overtaking scenario
were provided.
\vspace{-1mm}




%

\section*{ACKNOWLEDGMENT}
\vspace{-1mm}
The authors gratefully acknowledge 
Dmitry Yershov for insightful discussions.
\vspace{-2mm}


\bibliographystyle{IEEEtran}
\bibliography{IEEEabrv,bibs}

\end{document}